\documentclass[letterpaper]{article} % DO NOT CHANGE THIS
\usepackage{aaai24}  % DO NOT CHANGE THIS
\usepackage{times}  % DO NOT CHANGE THIS
\usepackage{helvet}  % DO NOT CHANGE THIS
\usepackage{courier}  % DO NOT CHANGE THIS
\usepackage[hyphens]{url}  % DO NOT CHANGE THIS
\usepackage{graphicx} % DO NOT CHANGE THIS
\usepackage{natbib}  % DO NOT CHANGE THIS AND DO NOT ADD ANY OPTIONS TO IT
\usepackage{caption} % DO NOT CHANGE THIS AND DO NOT ADD ANY OPTIONS TO IT
\usepackage{algorithm}
\usepackage{algorithmic}

\usepackage{newfloat}
\usepackage{listings}
\DeclareCaptionStyle{ruled}{labelfont=normalfont,labelsep=colon,strut=off} % DO NOT CHANGE THIS
\floatstyle{ruled}
\newfloat{listing}{tb}{lst}{}
\floatname{listing}{Listing}
\usepackage{arydshln} %%# 负责画虚线的包
\usepackage[table]{xcolor}
\usepackage{colortbl}
\usepackage{graphicx}
\usepackage{amsmath}
\usepackage{amssymb}
\usepackage{booktabs}
\usepackage{enumitem}
\usepackage{graphicx}
\usepackage{amsmath}
\usepackage{amssymb}
\usepackage{booktabs}
\usepackage{enumitem}
\usepackage{xcolor}
\usepackage{url}

\usepackage{booktabs, multirow} % for borders and merged ranges
\definecolor{roadcolor}{RGB}{234,51,246}
\definecolor{sidewalkcolor}{RGB}{68,8,72}
\definecolor{parkingcolor}{RGB}{241,156,249}
\definecolor{othergroundcolor}{RGB}{160,32,76}
\definecolor{buildingcolor}{RGB}{246,202,69}
\definecolor{carcolor}{RGB}{111,149,238}
\definecolor{truckcolor}{RGB}{74,32,172}
\definecolor{bicyclecolor}{RGB}{136,227,242}
\definecolor{motorcyclecolor}{RGB}{37,59,146}
\definecolor{othervehiclecolor}{RGB}{96,81,242}
\definecolor{vegetationcolor}{RGB}{79, 173, 50}
\definecolor{trunkcolor}{RGB}{126, 65, 22}
\definecolor{terraincolor}{RGB}{171, 238, 105}
\definecolor{personcolor}{RGB}{234, 60, 49}
\definecolor{bicyclistcolor}{RGB}{234, 66, 195}
\definecolor{motorcyclistcolor}{RGB}{138, 42, 90}
\definecolor{fencecolor}{RGB}{238, 128, 69}
\definecolor{polecolor}{RGB}{252, 241, 161}
\definecolor{trafficsigncolor}{RGB}{233, 51, 35}
\definecolor{color1}{RGB}{176, 36, 24}
\definecolor{color2}{RGB}{119,185,0}
\definecolor{color3}{RGB}{0, 0, 200}
\definecolor{colorofteaser}{RGB}{176, 36, 24}

\definecolor{LightGrey}{rgb}{.9,.9,.9}
\definecolor{White}{rgb}{1.,0.,1.}
\definecolor{first}{rgb}{.8,.0,.0}
\definecolor{second}{rgb}{.0,.6,.0}
\definecolor{third}{rgb}{.0,.0,.8}
\definecolor{ceiling}{RGB}{214,  38, 40}   %
\definecolor{floor}{RGB}{43, 160, 4}     %
\definecolor{wall}{RGB}{158, 216, 229}  %
\definecolor{window}{RGB}{114, 158, 206}  %
\definecolor{chair}{RGB}{204, 204, 91}   %
\definecolor{bed}{RGB}{255, 186, 119}  %
\definecolor{sofa}{RGB}{147, 102, 188}  %
\definecolor{table}{RGB}{30, 119, 181}   %
\definecolor{tvs}{RGB}{160, 188, 33}   %
\definecolor{furniture}{RGB}{255, 127, 12}  %
\definecolor{objects}{RGB}{196, 175, 214} %
\definecolor{car}{rgb}{0.39215686, 0.58823529, 0.96078431}
\definecolor{bicycle}{rgb}{0.39215686, 0.90196078, 0.96078431}
\definecolor{motorcycle}{rgb}{0.11764706, 0.23529412, 0.58823529}
\definecolor{truck}{rgb}{0.31372549, 0.11764706, 0.70588235}
\definecolor{other-vehicle}{rgb}{0.39215686, 0.31372549, 0.98039216}
\definecolor{person}{rgb}{1.        , 0.11764706, 0.11764706}
\definecolor{bicyclist}{rgb}{1.        , 0.15686275, 0.78431373}
\definecolor{motorcyclist}{rgb}{0.58823529, 0.11764706, 0.35294118}
\definecolor{road}{rgb}{1.        , 0.        , 1.        }
\definecolor{parking}{rgb}{1.        , 0.58823529, 1.        }
\definecolor{sidewalk}{rgb}{0.29411765, 0.        , 0.29411765}
\definecolor{other-ground}{rgb}{0.68627451, 0.        , 0.29411765}
\definecolor{building}{rgb}{1.        , 0.78431373, 0.        }
\definecolor{fence}{rgb}{1.        , 0.47058824, 0.19607843}
\definecolor{vegetation}{rgb}{0.        , 0.68627451, 0.        }
\definecolor{trunk}{rgb}{0.52941176, 0.23529412, 0.        }
\definecolor{terrain}{rgb}{0.58823529, 0.94117647, 0.31372549}
\definecolor{pole}{rgb}{1.        , 0.94117647, 0.58823529}
\definecolor{traffic-sign}{rgb}{1.        , 0.        , 0.    }   

\makeatletter
\newcommand{\car@semkitfreq}{3.92}
\newcommand{\bicycle@semkitfreq}{0.03}
\newcommand{\motorcycle@semkitfreq}{0.03}
\newcommand{\truck@semkitfreq}{0.16}
\newcommand{\othervehicle@semkitfreq}{0.20}
\newcommand{\person@semkitfreq}{0.07}
\newcommand{\bicyclist@semkitfreq}{0.07}
\newcommand{\motorcyclist@semkitfreq}{0.05}
\newcommand{\road@semkitfreq}{15.30}  %
\newcommand{\parking@semkitfreq}{1.12}
\newcommand{\sidewalk@semkitfreq}{11.13}  %
\newcommand{\otherground@semkitfreq}{0.56}
\newcommand{\building@semkitfreq}{14.1}  %
\newcommand{\fence@semkitfreq}{3.90}
\newcommand{\vegetation@semkitfreq}{39.3}  %
\newcommand{\trunk@semkitfreq}{0.51}
\newcommand{\terrain@semkitfreq}{9.17} %
\newcommand{\pole@semkitfreq}{0.29}
\newcommand{\trafficsign@semkitfreq}{0.08}
\newcommand{\semkitfreq}[1]{{\csname #1@semkitfreq\endcsname}}

\title{One at a Time: Progressive Multi-step Volumetric Probability Learning for \\Reliable 3D Scene Perception}
\author{
    Bohan Li\textsuperscript{\rm 1, 2},
    Yasheng Sun\textsuperscript{\rm 3},
    Jingxin Dong\textsuperscript{\rm 2},
    Zheng Zhu\textsuperscript{\rm 4},
    Jinming Liu\textsuperscript{\rm 1, 2},
    Xin Jin\textsuperscript{\rm 2}\thanks{Corresponding author.},
    Wenjun Zeng\textsuperscript{\rm 1, 2}
}
\affiliations{
    \textsuperscript{\rm 1}Shanghai Jiao Tong University, Shanghai, China\\
    \textsuperscript{\rm 2}Ningbo Institute of Digital Twin, Eastern Institute of Technology, Ningbo, China \\
    \textsuperscript{\rm 3}Tokyo Institute of Technology, Tokyo, Japan \\
    \textsuperscript{\rm 4}PhiGent Robotics, Beijing, China\\
    \{bohan\_li, jmliu206\}@sjtu.edu.cn, sun.y.aj@m.titech.ac.jp, \\
    jingxin\_dong@outlook.com, zhengzhu@ieee.org, \{jinxin, wenjunzengvp\}@eias.ac.cn
}

\usepackage{bibentry}
\begin{document}

\maketitle

\begin{figure}[H]
\hsize=\textwidth %
\begin{center}
\includegraphics[width=0.95\textwidth]{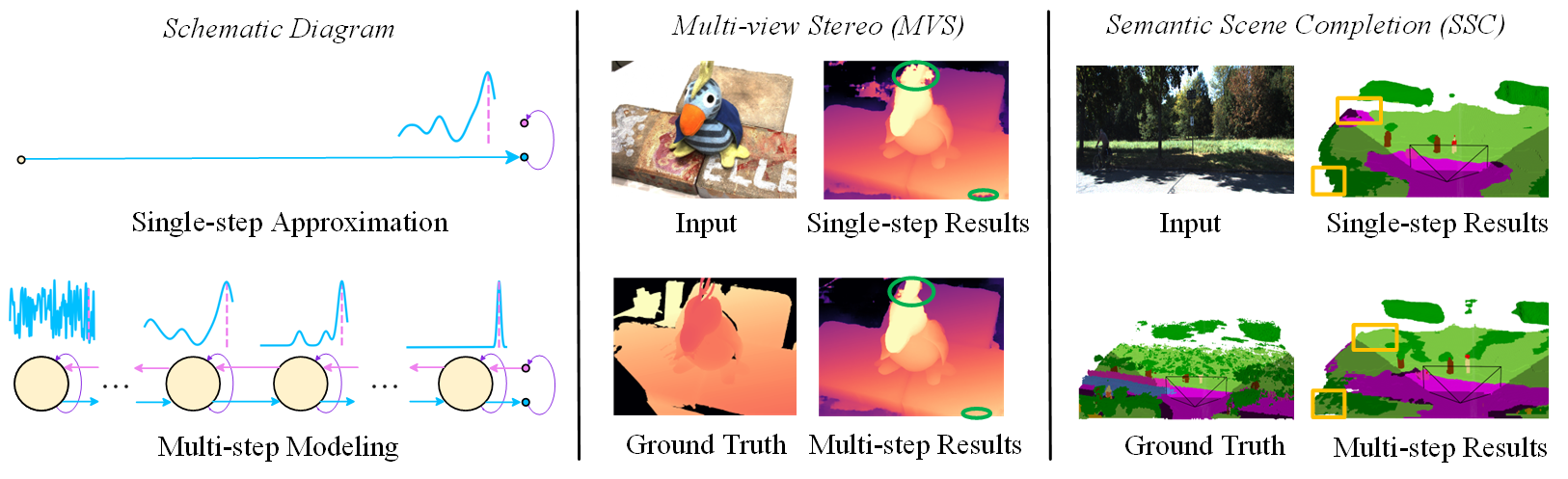}
\caption{
Comparison between single-step approximation and multi-step modeling for 3D scene perception tasks including multi-view stereo (MVS) and semantic scene completion (SSC). We demonstrate the qualitative results of these two methods. The multi-step modeling yields significantly more accurate and reliable results. } 
\label{teaser}
\end{center}
\end{figure}

\begin{abstract}
Numerous studies have investigated the pivotal role of reliable 3D volume representation in scene perception tasks, such as multi-view stereo (MVS) and semantic scene completion (SSC). They typically construct 3D probability volumes directly with geometric correspondence, attempting to fully address the scene perception tasks in a single forward pass. However, such a single-step solution makes it hard to learn accurate and convincing volumetric probability, especially in challenging regions like unexpected occlusions and complicated light reflections. Therefore, this paper proposes to decompose the complicated 3D volume representation learning into a sequence of generative steps to facilitate fine and reliable scene perception. Considering the recent advances achieved by strong generative diffusion models, we introduce a multi-step learning framework, dubbed as \textbf{VPD}, dedicated to progressively refining the \textbf{V}olumetric \textbf{P}robability in a \textbf{D}iffusion process. Specifically, we first build a coarse probability volume from input images with the off-the-shelf scene perception baselines, which is then conditioned as the basic geometry prior before being fed into a 3D diffusion UNet, to progressively achieve accurate 
\\
\\
\\
\\
\\
\\
\\
\\
\\
\\
\\
\\
\\
\\
\\
\\
\\
\\
\\
\\
\\
probability distribution modeling. To handle the corner cases in challenging areas, a Confidence-Aware Contextual Collaboration (CACC) module is developed to correct the uncertain regions for reliable volumetric learning based on multi-scale contextual contents. Moreover, an Online Filtering (OF) strategy is designed to maintain representation consistency for stable diffusion sampling. Extensive experiments are conducted on scene perception tasks including multi-view stereo (MVS) and semantic scene completion (SSC), to validate the efficacy of our method in learning reliable volumetric representations. Notably, for the SSC task, our work stands out as the first to surpass LiDAR-based methods on the SemanticKITTI dataset.
\end{abstract}

\section{Introduction}
Obtaining a dependable 3D representation is of critical importance in the realm of computer vision, particularly for tasks involving 3D scene perception, such as multi-view stereo (MVS)~\cite{yao2018mvsnet,chen2020mvsnet++,zhang2020adaptive} and semantic scene completion (SSC)~\cite{li2023voxformer,miao2023occdepth,li2023stereoscene}. 
The existing 2D-based approaches implicitly learned 3D features by harnessing contextual information~\cite{mayer2016large, wang2020fadnet, wang2021patchmatchnet}, which often struggle with precise geometric modeling due to the inherent ambiguity of 2D representations.
On the other hand, some researchers have sought to enforce geometric constraints by utilizing 3D probability volumes to model correspondences across various depth hypothesis planes, which attracts growing attention~\cite{yin2019hierarchical,gu2020cascade,ding2022transmvsnet}.

% Compared with these methods based on 2D contextual features~\cite{mayer2016large, wang2020fadnet, wang2021patchmatchnet}, the 3D volumetric-based approaches efficiently leverage stereo geometric constraints through the matching process and achieve compelling results across different benchmark datasets
% ~\cite{ ding2022transmvsnet, peng2022rethinking,xu2023iterative}. 

% \bh{DiffRF~\cite{muller2022diffrf} adopts a set of posed images as additional conditions for radiance field synthesis with a rendering loss to resolve ambiguities for 3D point clouds generation. }

% \bh{As shown in Figure~\ref{teaser}, }

Nevertheless, many complex real-world scenarios, characterized by incomplete observations and intricate reflection conditions, pose substantial challenges when striving for precise geometric modeling.
Existing 3D probability volume-based approaches have made strides by devising sophisticated architectures~\cite{gu2020cascade,chen2020mvsnet++,ding2022transmvsnet} and refining loss functions~\cite{peng2022rethinking,wang2022mvster} to acquire reliable probability volumes. However, these methods generally resolve the problem with a single-step approximation solution, imposing a substantially heavy burden on the learning process.
To mitigate these learning challenges, another line of research has introduced GRU-based architectures~\cite{yao2019recurrent,wang2022itermvs,xu2023iterative} to facilitate the acquisition of a dependable 3D volumetric representation through iterative refinement. 
Nevertheless, these approaches typically rely on 2D convolutional GRU mechanisms, which are susceptible to cumulative errors~\cite{li2017diffusion, mao2022review}.
This, in turn, motivates us to explore the potential of iterative refinement in the context of 3D volumetric probability.

Based on the above analysis, we propose a \textbf{Volumetric Probability Diffusion (VPD)} framework, which progressively models the volumetric probability and thus achieves reliable geometry estimation in the MVS and SSC tasks. As depicted in Fig.~\ref{teaser}, the core idea is to \emph{devise a multi-step learning scheme that models the probability volumes and progressively refine them}. 
Inspired by the powerful probability distribution modeling capabilities exhibited by generative diffusion models~\cite{saharia2022image,muller2022diffrf}, we propose a progressive optimization paradigm based on the diffusion process for reliable probability volume modeling.
% stable convergence. 
To leverage the geometry prior extracted from the input images with pre-trained models, our VPD is conditioned with the extracted coarse volumes and contextual features to guide the diffusion progress.
% As shown in Figure~\ref{teaser}, unlike the previous \textbf{single-step} approximation solutions, VPD takes the volumetric distribution approximation as a \textbf{multi-step} generative process of conditional probabilistic distribution, which progressively constructs the correspondence of target cost volumes with the conditional constraints. 
% Specifically, built upon the typical MVS~\cite{gu2020cascade,long2021multi,ding2022transmvsnet,peng2022rethinking} or SSC~\cite{li2023stereoscene} baselines, VPD innovatively constructs two kinds of conditional guidance to lead the distribution transition, including probability volumes via \textbf{Conditional Volume Probabilization (CVP)} and confidence-aware context via a \textbf{Confidence-Aware Contextual Collaboration (CACC)} module. 
% Particularly, {CVP means we extract the coarse cost volumes from the pre-trained baselines as basic prior knowledge and transform them into probabilistic form \bh{same as typical MVS pipelines\cite{yao2018mvsnet,gu2020cascade} to condition the diffusion model for target distribution approximation.}
% Despite the advance of CVP in high-confidence regions,
Specifically, the coarse volumes are employed as basic geometry prior, which is concatenated with the noisy input volume of the diffusion framework as prior volume
condition. 
Despite the effectiveness of the prior volume condition in high-confidence regions, the low-confidence mismatch issue in challenging regions (e.g. non-Lambertian surfaces, thin structures and reflections) still exists, which impairs the learning of probability distribution approximation to the target volumes. Therefore, we further introduce a \textbf{Confidence-Aware Contextual Collaboration (CACC)} module to correct the uncertain regions of the predicted 3D volumes with rich contextual information. In detail, CACC first prunes the 3D volumes using confidence-aware filtering. Next, the fine-grained features and geometric details are retrieved from multi-scale contextual contents to complement the information in the low-confidence regions of the volumes. 
Moreover, to avoid perturbations in the diffusion sampling process, we introduce an \textbf{Online Filtering (OF)} strategy to maintain the consistency of the representations for a stable diffusion. In summary, the main contributions of this paper are listed as:

\begin{itemize}

\item We pinpoint the limitation of sing-step-based strategies, and correspondingly propose a novel Volumetric Probability Diffusion (VPD) framework, which fully exploits the strong generative ability of diffusion models for fine and reliable volumetric representation.
% which progressively learns a reliable volumetric probability in a multi-step manner\bh{with a strong generative diffusion process}.
% models complex depth volumes using multi-step generative diffusion to obtain a final accurate prediction. 
% The probability volumes generated from Coarse Volume Probabilization (CVP) are adopted as basic prior knowledge.
\item  We propose a Confidence-Aware Contextual Collaboration (CACC) module to enhance the reliability of volumetric learning in VPD. Additionally, we develop an Online Filtering (OF) strategy to maintain representation consistency during the reverse sampling process.
\item 
Extensive experiments validate the effectiveness of our approach. We achieve state-of-the-art results on various scene perception tasks, including 1) MVS: DTU~\cite{article}, BlendedMVS~\cite{yao2020blendedmvs} and ScanNet~\cite{dai2017scannet}; 2) SSC: SemanticKITTI~\cite{behley2019semantickitti}. Notably, to the best of our knowledge, VPD is the first camera-based method that surpasses LiDAR-based methods on the SemanticKITTI.
\end{itemize}

% We evaluate our VPD on various popular benchmark datasets of multi-view stereo and semantic scene completion. For multi-view stereo, our method significantly surpasses mainstream methods on DTU~\cite{}, BlendedMVs~\cite{} and ScanNet~\cite{} datasets. 
% For semantic scene completion, our VPD achieves state-of-the-art performance on SemanticKITTI~\cite{} when built upon StereoScne~\cite{}. 

\section{Related Works}

\subsection{Learning-based 3D Scene Perception}
With the development of learning-based methods, the quality of  3D representation for scene perception has been steadily improved~\cite{Lahav2021raftstereo,okae2021robust,xie2023navinerf}.
% including stereo matching, semantic scene completion, etc. 
% Broadly, stereo matching can be divided into binocular stereo and multi-view stereo depending on the number of input images~\cite{gu2020cascade,poggi2022stereodepth,xu2023iterative}. Binocular stereo uses rectified stereo images as input and 
Recently, stereo matching has been explored in semantic scene completion (SSC)~\cite{li2023voxformer,li2023stereoscene}. 
% Voxformer~\cite{li2023voxformer} employs a two-stage framework where depth estimation is utilized to construct voxel queries in the first stage, and further guide dense voxel generation in the following stage.
In StereoScene~\cite{li2023stereoscene}, a stereo volume constructor is proposed to generate a geometric cost volume to enhance the understanding of 3D scenarios.
For multi-view stereo (MVS), a set of images are employed to construct 3D cost volumes with epipolar constrain~\cite{gu2020cascade,ding2022transmvsnet,peng2022rethinking}. 
% MVSNet~\cite{yao2018mvsnet} proposes to build variance-based cost volumes to measure similarity among multiple views. 
% ESTD~\cite{long2021multi} constructs geometric correlations with temporal coherence using an epipolar spatio-temporal transformer for multi-view depth estimation. 
CasMVSNet~\cite{gu2020cascade} employs cascade cost volumes with different scales to form a coarse-to-fine depth estimation framework. TransMVSNet~\cite{ding2022transmvsnet} leverages global context information with a feature matching transformer to exploit long-range aggregation across input images. UniMVSNet~\cite{peng2022rethinking} proposes a unification representation for both regression and classification that is supervised with a unified focal loss. Different from all the previous methods that try to approximate the ground truth in a single step, we propose to formulate depth estimation as progressive distribution modeling, which decomposes the issue into multiple steps to further improve performance in challenging scenarios.

% \subsection{GRU-based Iterative Optimization }
% \bh{
% Several depth estimation methods employ GRU-based iterative schemes for performance improvement~\cite{yao2019recurrent,wang2022itermvs,xu2023iterative}.
% R-MVSNet~\cite{yao2019recurrent} proposes to view the 3D cost volume as multiple 2D cost maps, and utilizes 2D convolutional GRU to aggregate temporal context information along the depth dimension. Consequently, the estimation results with GRU-based R-MVSNet are inevitably affected by previous steps, leading to temporal cumulative errors~\cite{li2017diffusion, mao2022review}. On the other hand, our proposed method could achieve more robust predictions by considering the 3D cost volume as input in every iterative step.
% Following R-MVSNet, more GRU-based methods including IterMVS~\cite{wang2022itermvs}, RAFT-Stereo~\cite{Lahav2021raftstereo}, IGEV-Stereo~\cite{xu2023iterative} etc., employ direct supervision between the predictions and the ground truth, which essentially forces the estimated distribution to approximate the complex target distribution in a single step. Different from that, our method constructs intermediate target distribution at each step to subtly split the learning into multi-step modeling.}

\begin{figure*}[!ht]
\hsize=\textwidth %
\centering
\includegraphics[width=0.95\textwidth]{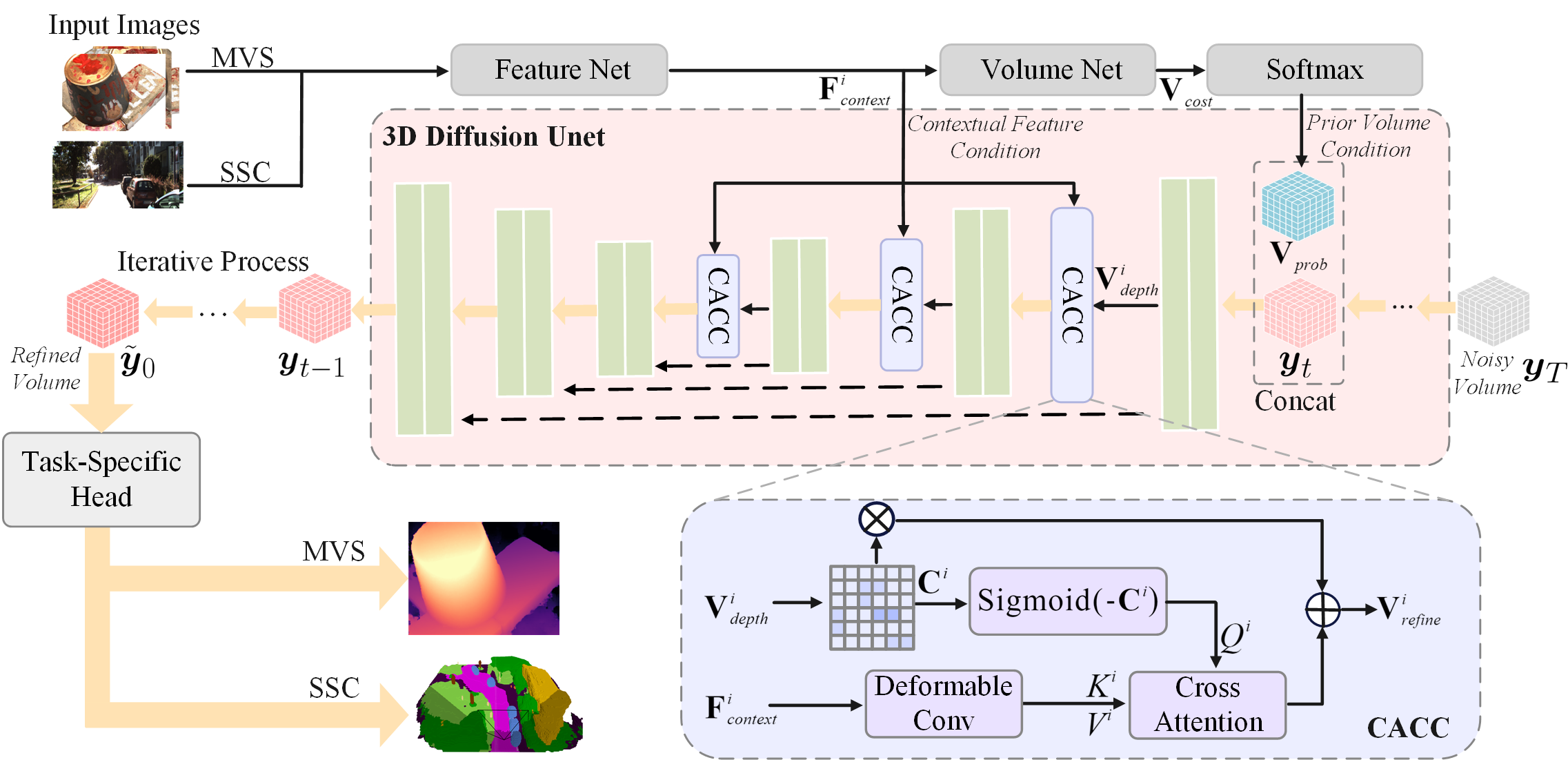}
\caption{Overall framework of the proposed volumetric probability diffusion (VPD). 
Given input images, We first extract multi-scale contextual features $\textbf{F}_{context}^{i}$ and coarse probabilistic volumes $\textbf{V}_{prob}$ with off-the-shelf scene perception baselines. 
Then, $\textbf{V}_{prob}$ concatenated with the random noisy volume ${\boldsymbol{y}}_t$ as input is fed into the 3D diffusion UNet for refinement, while $\textbf{F}_{context}^{i}$ are employed as conditions in CACC to continuously refine the depth volume $\textbf{V}_{depth}^{i}$ in the 3D UNet. Following an iterative process, we progressively estimate a refined volume $\tilde{\boldsymbol{y}}_0$ over multiple steps with diffusion. The estimated volumes are finally fed to the task-specific head to generate depth maps for MVS or occupancy grids for SSC.}
\label{overall}
\end{figure*}

\subsection{ Denoising Diffusion Models }
Denoising diffusion models (DDMs) are a novel class of generative models derived from nonequilibrium thermodynamics~\cite{sohl2015deep} and have achieved astounding results in the field of computer vision~\cite{luo2021diffusion, rombach2022high,ramesh2021zero}. 
% natural language processing~\cite{li2022diffusion, gong2022diffuseq, reid2022diffuser} and AI for science~\cite{luo2021predicting, xu2022geodiff}. 
% DDMs model empirical data distribution by iterative denoising process~\cite{shao2022diffustereo} and closely resemble score-based generative models, which generate samples by Langevin dynamics based on estimated gradients of the data distribution~\cite{jiang2022conditional}. %The vanilla diffusion models can be used for prediction tasks by designing additional conditions as guidance. 
% Controlling the behavior of the vanilla diffusion models can optimize results on prediction tasks by designing additional conditions as guidance. For example, 
% SR3~\cite{saharia2022image} produces photo-realistic samples on super-resolution tasks based on low-resolution images as the condition. 
DiffRF~\cite{muller2022diffrf} adopts a set of posed images as additional conditions for radiance field synthesis with a rendering loss to resolve ambiguities.
{Different from the one-to-many mapping in the generation process of DiffRF, we employ the diffusion process as a one-to-one mapping, leveraging geometry prior for accurate and reliable 3D scene perception.}
DiffRF~\cite{muller2022diffrf} adopts a set of posed images as additional conditions for radiance field synthesis with a rendering loss to resolve ambiguities.
{Different from the one-to-many mapping in the generation process of DiffRF, we employ the diffusion process as a one-to-one mapping, leveraging geometry prior for accurate and reliable 3D scene perception.}
DiffuStereo~\cite{shao2022diffustereo} leverages an iterative diffusion model to obtain highly accurate depth maps for automatic high-quality human reconstruction from sparse-view inputs as conditions. However, DiffuStereo directly refines the depth maps generated by the off-the-shelf algorithms, without fully exploring the geometric constraints in the matching process. In contrast, we propose volumetric probability diffusion (VPD) to make full use of the correspondence distribution across different depth hypothesis planes, which is more advisable because the diffusion process excels at modeling distributions.

% we construct conditions for the diffusion process with off-the-shelf scene perception baselines. 

% diffusing a noisy input volume ${\boldsymbol{y}}_T$ with the conditional guidance. 

\section{Methodology}

In this work, we formulate the 3D perception in MVS and SSC tasks as multi-step conditional volumetric probability learning, and propose Volumetric Probability Diffusion (VPD).
As shown in Figure~\ref{overall}, given input images, we first construct diffusion conditions with coarse probabilistic volumes and multi-scale contextual features extracted from off-the-shelf scene perception baselines. 
Next, we progressively estimate a refined volume over multiple steps by diffusing a noisy volume with the constructed conditions.
The refined volumes are finally fed into the task-specific head to generate depth maps in MVS or occupancy grids in SSC. Please refer to the \textbf{Supplementary Material} for the details on the task-specific head.
% To this end, we introduce VPD to learn a parametric approximation to the task-specific distribution progressively. 
% As shown in Figure~\ref{overall}, VPD predicts target volumes taking noisy volumes with two conditions as guidance. The conditions are constructed from multi-view images based on MVS or SSC baselines, while the estimated volumes are fed to the task-specific head to generate depth maps in MVS or occupancy grids in SSC. 
In detail, the proposed VPD mainly consists of the following components: \looseness=-1

\textbf{\uppercase\expandafter{\romannumeral1.}} A \textbf{Volumetric Diffusion} model that is implemented with a 3D UNet~\cite{ronneberger2015u}.  
In the forward process, the target volumes are constructed from ground truth depth maps. In the reverse process, an Online Filtering (\textbf{OF}) strategy is further developed to maintain the
unique peak distribution in the estimated volumes.

\textbf{\uppercase\expandafter{\romannumeral2.}} The diffusion conditions including the basic prior volume condition and the contextual feature condition constructed with the Confidence-Aware Contextual Collaboration (\textbf{CACC}) module.

% To condition the diffusion process in VPD, the coarse probability volumes are constructed with the off-the-shelf scene perception baselines as the basic prior volume condition. 
% Moreover, the Confidence-Aware Contextual Collaboration (\textbf{CACC}) module is proposed to construct the contextual feature condition to further refine the estimated volumes in the 3D diffusion UNet.

% \begin{algorithm}[H]
% \KwIn{$x$, $y$}
% \KwOut{$x$}
% \BlankLine
% \caption{ Training a denoising model $f_\theta$} \label{}
% \Repeat{converged}{
%     \State $(x, y_0) \sim p(x, y)$\;
%       \State $\gamma \sim p(\gamma)$\;
      
%       \State $\bm{\epsilon}\sim\mathcal{N}(\mathbf{0},\mathbf{I})$\;
      
%       \State Take a gradient descent step on\;
%             \Statex $ \quad \nabla_\theta\left\|f_\theta\left(x, \sqrt{\gamma} y_0+\sqrt{1-\gamma} \epsilon, \gamma\right)-\epsilon\right\|_p^p $\;

% }

% % \textbf{output} $x$\;
% \end{algorithm}

% \begin{algorithm}[H]
% \KwIn{  $  y_T \sim \mathcal{N}(\mathbf{0}, \mathbf{I})  $\; }
% \KwOut{$\boldsymbol{y}_{0}$}
% \BlankLine
% \caption{ Inference in refinement steps } \label{}
% \For{$t=T, \dotsc, 1$}{
    
%        $z \sim \mathcal{N}(\mathbf{0}, \mathbf{I})$ if $t > 1$, else $z = \mathbf{0}$\;
       
%        $y_{t-1} = \frac{1}{\sqrt{\alpha_t}}\left( y_t - \frac{1-\alpha_t}{\sqrt{1-\gamma_t}} f_\theta(x, y_t, \gamma_t) \right) + \sqrt{1 - \alpha_t} z$\;
%        }

% \textbf{return} $\boldsymbol{y}_{0}$\;
% \end{algorithm}

\subsection {Volumetric Diffusion}\label{sec31}

The standard generative diffusion models aim to form one-to-many mappings with a forward and reverse process.
In our scenario, we employ a volumetric diffusion model to learn the parametric approximation to the target volume based on the guidance of conditions. 

In the forward process, we construct target unimodel volume $\boldsymbol{y}_{0}$ from ground truth depth map $d^{gt}$, and progressively corrupt the target volume to $\boldsymbol{y}_{T} \sim \mathcal{N}(\mathbf{0}, \boldsymbol{I})$ in $\boldsymbol{T}$ time steps. 
In the reverse process, the 3D diffusion UNet estimates a refined volume $\tilde{\boldsymbol{y}}_0$ to approximate the target volume $\boldsymbol{y}_{0}$ from noisy input volume $\boldsymbol{y}_{T}$ and we consider conditions $\boldsymbol{x}$ to guide the estimation. %$f_\theta$

\subsubsection{Volumetric Gaussian Forward Process.} \label{sec311}

Given a ground truth depth map $d^{gt}$, we first construct the target volume $\boldsymbol{y}_{0}$ following the unimodel projection~\cite{ding2022transmvsnet, peng2022rethinking} along depth dimension $D$ as diffusion input:
 \begin{equation}
\boldsymbol{y}_0= Project^{Uni} \left\{ d^{gt}, Dim=D \right\}, 
 \end{equation}
 
We gradually add noise on $\boldsymbol{y}_{0}$ to generate the noisy volume $\boldsymbol{y}_{T}$ over $\boldsymbol{T}$ steps following a discrete-time Markov chain.
% \begin{equation}
% q\left(\boldsymbol{y}_{1: T} \mid \boldsymbol{y}_0\right)=\prod_{t=1}^T q\left(\boldsymbol{y}_t \mid \boldsymbol{y}_{t-1}\right)
% \end{equation}
% \begin{equation}
% q\left(\boldsymbol{y}_t \mid \boldsymbol{y}_{t-1}\right)=\mathcal{N}\left(\boldsymbol{y}_t \mid \sqrt{\alpha_t} \boldsymbol{y}_{t-1},\left(1-\alpha_t\right) \boldsymbol{I}\right)
% \end{equation}
% where the hyper-parameter $\alpha_t$ dominates injected noise variance at each iteration. 
% $\mathcal{N}$ and $\boldsymbol{I}$ denote multivariate normal distribution and identical matrix, respectively. 
% The $\sqrt{\alpha_t}$ agradually attenuates with $\boldsymbol{y}_{t-1}$, making $\boldsymbol{y}_{T}$ typically close to the complete random noisy volume according to predefined noise-variance schedule. 
Given distribution of $\boldsymbol{y}_{0}$, the forward process can be characterized as:
\begin{equation}
q\left(\boldsymbol{y}_t \mid \boldsymbol{y}_0\right)=\mathcal{N}\left(\boldsymbol{y}_t\mid\sqrt{\bar{\alpha}_t} \boldsymbol{y}_0,\left(1-\bar{\alpha}_t\right) \boldsymbol{I}\right) ,
\end{equation}
where $\bar{\alpha}_t=\prod_{i=1}^t \alpha_i$ and $\alpha_t$ is the pre-defined coefficient. $\mathcal{N}$ and $\boldsymbol{I}$ denote the normal distribution and the identical matrix, respectively.

% The posterior distribution of ${y}_{t-1}$ can be further derived 
% % by exploiting algebraic manipulation 
% with $({y}_{0},{y}_{t})$:
% \begin{equation}
%  q\left(\boldsymbol{y}_{t-1} \mid \boldsymbol{y}_0, \boldsymbol{y}_t\right)=\mathcal{N}\left(\boldsymbol{y}_{t-1} \mid \boldsymbol{\mu}, \sigma^2 \boldsymbol{I}\right) 
% \end{equation}
% where $\boldsymbol{\mu}=\frac{\sqrt{\bar{\alpha}_{t-1}}\left(1-\alpha_t\right)}{1-\bar{\alpha}_t} \boldsymbol{y}_0+\frac{\sqrt{\alpha_t}\left(1-\bar{\alpha}_{t-1}\right)}{1-\bar{\alpha}_t} \boldsymbol{y}_t $, and 
% $\sigma^2=\frac{\left(1-\bar{\alpha}_{t-1}\right)\left(1-\alpha_t\right)}{1-\bar{\alpha}_t}$. 

 \subsubsection{Iterative Conditional Reverse Process.}
%  Formally, the inferring process of our framework can be defined on the conditional distribution $\boldsymbol{p}_{\theta}(\boldsymbol{y}_0|\boldsymbol{x})$:
% \begin{equation}\label{}
% p_{\theta}\left(\boldsymbol{y}_{0: T} \mid \boldsymbol{x} \right)=p\left(\boldsymbol{y}_{T}\right) \prod_{t=1}^{T} p_{\theta}\left(\boldsymbol{y}_{t-1} \mid \boldsymbol{y}_{t}, \boldsymbol{x}  \right)
% \end{equation}

 The conditional reverse sampling process is dedicated to iteratively denoise $\boldsymbol{y}_{T}$ with conditions to recover $\boldsymbol{y}_{0}$.
 Each step of the reverse process can be defined as conditional distribution transition~\cite{saharia2022image}, which is formulated as:
 \begin{equation}
 p_\theta\left(\boldsymbol{y}_{0: T} \mid \boldsymbol{x}\right)=p\left(\boldsymbol{y}_T\right) \prod_{t=1}^{T} p_\theta\left(\boldsymbol{y}_{t-1} \mid \boldsymbol{y}_t, \boldsymbol{x}\right),
 % p_\theta\left(\boldsymbol{y}_{t-1} \mid \boldsymbol{y}_t, \boldsymbol{x}\right)=\mathcal{N}\left(\boldsymbol{y}_{t-1} \mid \mu_\theta\left( \boldsymbol{y}_t, \bar{\alpha}_t, \boldsymbol{x} \right), \sigma_t^2 \boldsymbol{I}\right) ,
\end{equation}
where $p_\theta$ represents the reverse function, $\boldsymbol{x}$ denotes the two conditions of the diffusion model. 
% $\mu_\theta\left( \boldsymbol{y}_t, \bar{\alpha}_t, \boldsymbol{x} \right)=\frac{1}{\sqrt{\alpha_t}}\left(\boldsymbol{y}_t-\frac{1-\alpha_t}{\sqrt{1-\bar{\alpha}_t}} f_\theta\left( \boldsymbol{y}_t, \bar{\alpha}_t,\boldsymbol{x} \right)\right)$, and 
% $\sigma^2=\frac{\left(1-\bar{\alpha}_{t-1}\right)\left(1-\alpha_t\right)}{1-\bar{\alpha}_t}$. 

\emph{Online Filtering.} %跟前向过程每一步构建的GT做对应，每一步的GT就是个单峰分布+高斯噪声。我们滤波之后只保留单峰分布，然后加上高斯噪声送到下一步采样中 intermediate noisy target distributions at each step
Since VPD is dedicated to learning for approximating the target volume $\boldsymbol{y}_0$ with unimodal distribution, we propose to filter the predicted ${\boldsymbol{y}}_t$ online at each iteration to suppress the perturbation caused by the generated multi-model representation before sending it to the next reverse sampling step. Our implementation can be formally written as:
 \begin{equation}
{\boldsymbol{y}}_t'= Project^{Uni} \left\{ WTA^{D} \left (  {\boldsymbol{y}}_t \right ), Dim=D \right\}, 
 \end{equation}
 where $Project^{Uni}$ denotes unimodel projection same as GT volume construction in Section~\ref{sec311}. $WTA^{D}$ represents \textit{Winner-Takes-All} operation~\cite{cheng2020hierarchical}, which maintains the unique peak along the depth dimension.

\subsection{ Condition Construction } 
In this section, we introduce the condition construction in VPD. As shown in Figure~\ref{overall}, we extract coarse probabilistic volumes and multi-scale contextual features from input images with the off-the-shelf MVS or SSC baselines (correspond to the Feature Net and the Volume Net in Figure~\ref{overall}). Next, we employ them as the prior volume condition and the contextual feature condition to constrain the learning of distribution transition in the diffusion process, respectively.

\subsubsection{Coarse Volume Probabilization.}\label{sec321}
We employ Coarse Volume Probabilization (CVP) to 
construct the coarse probabilistic volume, which is concatenated with the input noisy volume as the basic prior volume condition of the diffusion model.
% for a reasonable approximation in high-confidence regions. 
Given a coarse cost volume $\textbf{V}_{cost}$ from baseline networks, we employ $softmax$ along the depth dimension for each pixel $(h,w)$ in space to implement the volume probabilization, which is formally written as:
\begin{equation}
{\textbf{V}_{prob}^{h,w,m} = Softmax (\textbf{V}_{cost}^{h,w,m}) =   \frac{\exp(d_{m}^{h,w})} 
 {\sum_{n=1}^{D_{max}}\exp(d_{n}^{h,w})}  ,}
\end{equation}
where $d_{m}^{h,w}$ represents cost value of $m^{th}$ depth hypothesis plane ($1<m<D_{max}$). $D_{max}$ denotes the number of depth hypothesis planes. 
For multi-view stereo (MVS), we adopt regularized cost volumes~\cite{gu2020cascade,long2021multi,ding2022transmvsnet,peng2022rethinking} as the volume $\textbf{V}_{cost}$, while the geometric cost volumes~\cite{li2023stereoscene} are employed for semantic scene completion (SSC).
% Please refer to Section~\ref{sec43} for more details about the selection of the probabilization strategies. 

\begin{figure}[!ht]
\centering
\includegraphics[width=0.99\linewidth]{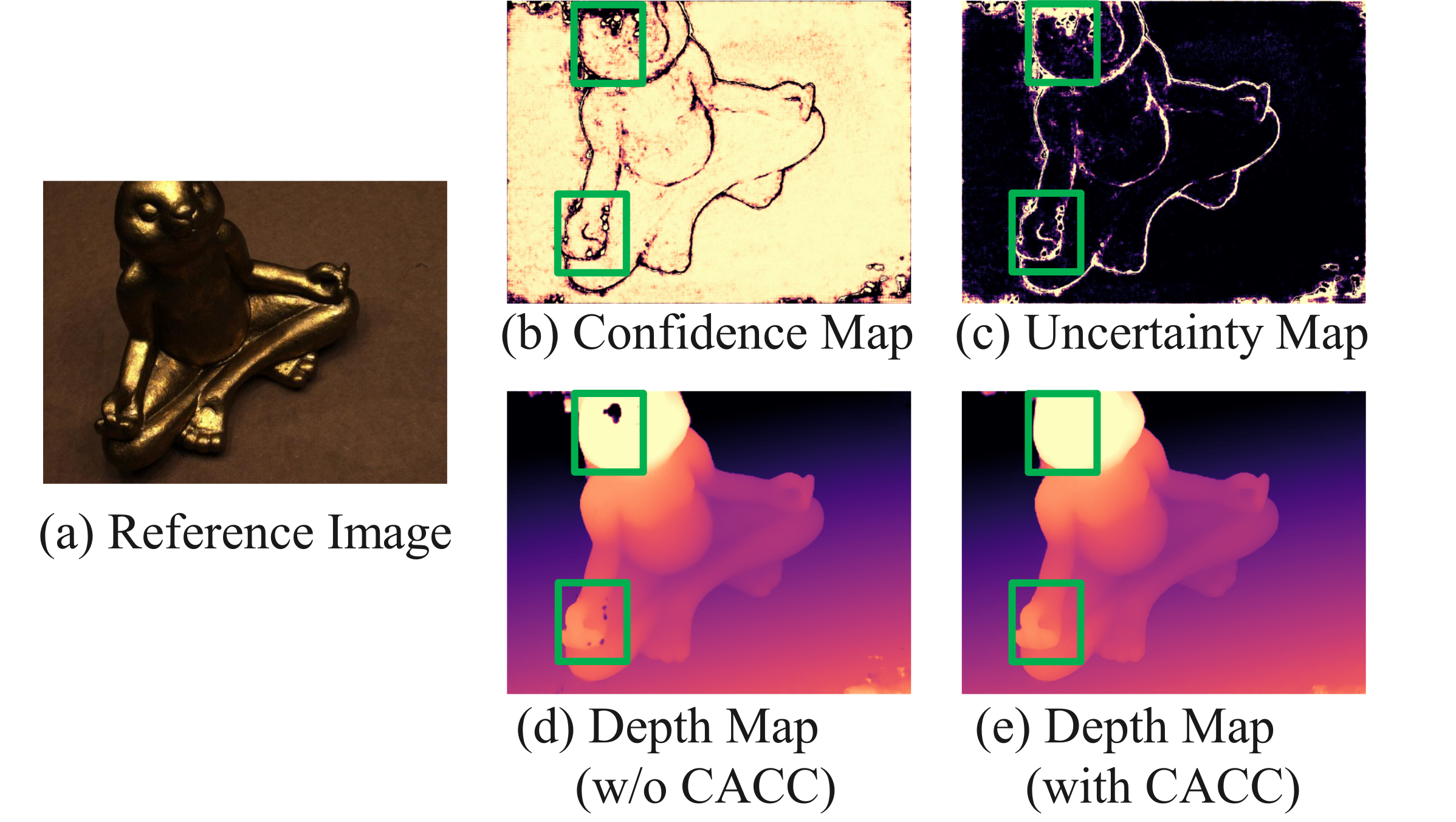}
\caption{Visualization results in the confidence-aware contextual collaboration (CACC) module. The confidence map and the uncertainty map illustrate the regions with poor estimation, which are effectively refined with CACC.}
\label{confidence}
\end{figure}

\subsubsection{Confidence-Aware Contextual Collaboration.}
Although the CVP provides basic geometry prior, it is still hard to achieve compelling results, especially in challenging regions like occlusions, reflections, textureless regions, etc. Thus, we propose a Confidence-Aware Contextual Collaboration (CACC) module to further apply continuous refinement with the contextual feature condition on the estimated volumes.
 
The overall structure of CACC is shown in Figure~\ref{overall}. 
Given a depth volume $\textbf{V}_{depth}^{i}\in \mathbb{R}^{C\times D\times H\times W}$ in $i^{th}$ downsample block of the 3D UNet (i.e., diffusion model) and $i^{th}$ scale contextual features $\textbf{F}_{context}^{i}\in \mathbb{R}^{C^{\prime}\times H\times W}$ from feature extraction networks, our goal is to retrieval reliable multi-scale contextual features from $\textbf{F}_{context}^{i}$, and refine $\textbf{V}_{depth}^{i}$ according to the confidence information along the spatial dimension. It is worth noting that we directly obtain multi-scale contextual features from the off-the-shelf baseline networks for computational efficiency.

Specifically, we first form a confidence map $\textbf{C}^{i}\in \mathbb{R}^{C\times H\times W}$ by checking the highest probability value among all depth hypothesis planes across the depth dimension. Next, we reverse the values in $\textbf{C}^{i}$ to obtain query $Q^{i}$ for cross attention that measures the matching uncertainty in $\textbf{V}_{depth}^{i}$:
\begin{equation}
    Q^{i}  = Sigmoid(-{\mathbf{C}}^{i}) =Sigmoid\left\{ - \left ( WTA^{D}\left ( \mathbf{V}^{i} \right )  \right ) \right\},
\end{equation}
where $WTA^{D}$ denotes \textit{Winner-Takes-All} operation along depth dimension. To generate key $K^{i}$ and value $V^{i}$, we apply deformable convolution on the corresponding contextual features $\textbf{F}_{context}^{i}$ for efficient geometric transformation modeling and receptive fields adaption. For each location point $\textbf{p}$ on the contextual features $\textbf{F}_{context}^{i}$, the process is formulated as:
\begin{equation}  
K^{i}(\textbf{p}), V^{i}(\textbf{p}) = Chunk \left\{ \sum_{c=0}^{C^{\prime}-1}  W \cdot \textbf{F}_{context}^{i}\left((\textbf{p})+\Delta_{(\textbf{p})},\mathbf{c} \right)
% , Dim=1
\right\},
\end{equation}
where $W$ and $\Delta_{(\textbf{p})}$ denotes the deformable weight and learnable offset, respectively. {$Chunk$ represents splitting the input into halves along the feature channel. }
To reduce computation cost, we adopt linear attention~\cite{kitaev2020reformer,shen2021efficient} as:
\begin{equation}
\textbf{F}_{conf}^{i}=LinearAtten(Q^{i}, K^{i}, V^{i})=\phi_{q}(Q^{i}) \left(  \phi_{k} (K^{i})^{T} V^{i} \right ),
\end{equation}
where $\textbf{F}_{conf}^{i}$ represents confidence-aware context. $\phi_{q}$ and $\phi_{k}$ are $softmax$ operations along each row and column of the input matrix, respectively. 
In this way, the relevant information of contextual features is retrieved according to the matching uncertainty of the depth volume $\textbf{V}_{depth}^{i}$.

Subsequently, we implement element-wise multiplication between the depth volume $\textbf{V}_{depth}^{i}$ and confidence map $\textbf{C}^{i}$ to obtain a filtered volume. 
To match in dimension, $\textbf{F}_{conf}^{i}$ is projected into 3D contextual volume $\textbf{V}_{context}^{i}$ before adding to the filtered volume following lift operation~\cite{philion2020lift}.
Finally, the refined volume is constructed by element-wise summation between the filtered volume and the contextual volume:
\begin{equation}
\textbf{V}_{refine}^{i} = \textbf{V}_{depth}^{i}  \odot \textbf{C}^{i} + \textbf{V}_{context}^{i},
\end{equation}
where $\odot$ denotes element-wise multiplication. Note that CACC is applied on each downsample block of the 3D UNet with different dimension sizes.
Through the refinement operation of CACC on the depth volume, volumetric distribution in high-confidence regions is retained, while that in low-confidence regions is optimized with multi-scale contexts.

In Figure~\ref{confidence}, we visualize the confidence map $\textbf{C}^{i}$, uncertainty map $Sigmoid(-{\mathbf{C}}^{i})$, estimated depth map without CACC and estimated depth map with CACC. It can be seen that the model without CACC struggles to achieve compelling results in challenging regions (e.g. object boundaries, low-texture regions).
The confidence map and the uncertainty map illustrate the regions with poor estimation, which are effectively refined by retrieving information from the contextual features with CACC.\looseness=-1

\subsection{ Training Objective }

In this work, we adopt an end-to-end joint training pipeline for the whole framework, and our training objective is to optimize the volumetric diffusion model for target volume approximation.
Different loss functions are applied to achieve the object according to the representations of coarse probabilistic volumes (in Section~\ref{sec321}), which is consistent with baseline networks~\cite{gu2020cascade,peng2022rethinking,ding2022transmvsnet,long2021multi,li2023stereoscene}. 
% Please refer to the \textbf{Supplementary Material} for the details on baseline loss functions.

\textbf{Regression Loss.} For regression representation, we implement implicit supervision on the output of VPD. Specifically, the estimated probabilistic depth volume is first regressed into a 2D depth map $\tilde{d}$, then a $Smooth L_1$ loss~\cite{gu2020cascade} is implemented between the estimated depth map $\tilde{d}$ and the ground truth $d^{gt}$:
\begin{equation}
\mathcal{L}_{Regress}=\frac{1}{N} \sum_{i=1}^N \operatorname{smooth}_{L_1}\left(\tilde{d}_{i}-d^{gt}_{i}\right) ,
\end{equation}
where $N$ denotes the number of labled pixels $d^{gt}_{i}$.
In this way, the depth volume predicted by the volumetric diffusion model is implicitly supervised throughout the training process.\looseness=-1

\textbf{Classification Loss.} For classification representation, we apply focal loss~\cite{ding2022transmvsnet} to directly supervise discrete volumetric distribution. The function adopts a tunable parameter $\gamma$ to help focus on hard samples to prevent overfitting, which is formally defined as:
\begin{equation}
\mathcal{L}_{Classify}=\sum_{\mathbf{p} \in\left\{\mathbf{p}_{v}\right\}}-\left(1-P^{(\tilde{\mathbf{h}})}(\mathbf{p})\right)^{\gamma} \log \left(P^{(\tilde{\mathbf{h}})}(\mathbf{p})\right)  ,
\end{equation}
where $\mathbf{p}_{v}$ and $\tilde{\mathbf{h}}$ denotes labeled pixels with valid ground truth and the depth hypothesis closest to the ground truth, respectively.
$P^{(\tilde{\mathbf{h}})}(\mathbf{p})$ represents the prediction on depth hypothesis $\tilde{\mathbf{h}}$.

\textbf{Unification Loss.} For unification representation, we adopt unified focal loss~\cite{peng2022rethinking} for continuous supervision on the estimated volume:
\begin{equation}
{ \mathcal{L}_{Unify}=  \begin{cases}\alpha^{+}(S_{b}^{+}(\frac{|q-u|}{q^{+}}))^{\gamma}BCE(u,q),  q>0\\\alpha^{-}(S_{b}^{-}(\frac{u}{q^{+}}) )^{\gamma}BCE(u,q),  else\end{cases} }.
\end{equation}
where $u$ and $q$ denote the estimated unity value and the continuous target, respectively. $\alpha$ and $\gamma$ are tunable parameters for sample balance.
$BCE$ represents binary cross-entropy. $S_{b}(x)$ is a sigmoid-like function as $(1/(1+b^{-x}))$.
Moreover, the $\alpha^{+}$ and $\alpha^{-}$ represent $\alpha$ and $1-\alpha$, respectively. The $S_{b}^{+}$ and $S_{b}^{-}$ represent the same sigmoid-like function with different inputs of $\frac{|q-u|}{q^{+}}$ and $\frac{u}{q^{+}}$, respectively. The $q^{+}$ denotes the positive target.

\begin{figure}[!ht]
\centering
\includegraphics[width=1.0\linewidth]{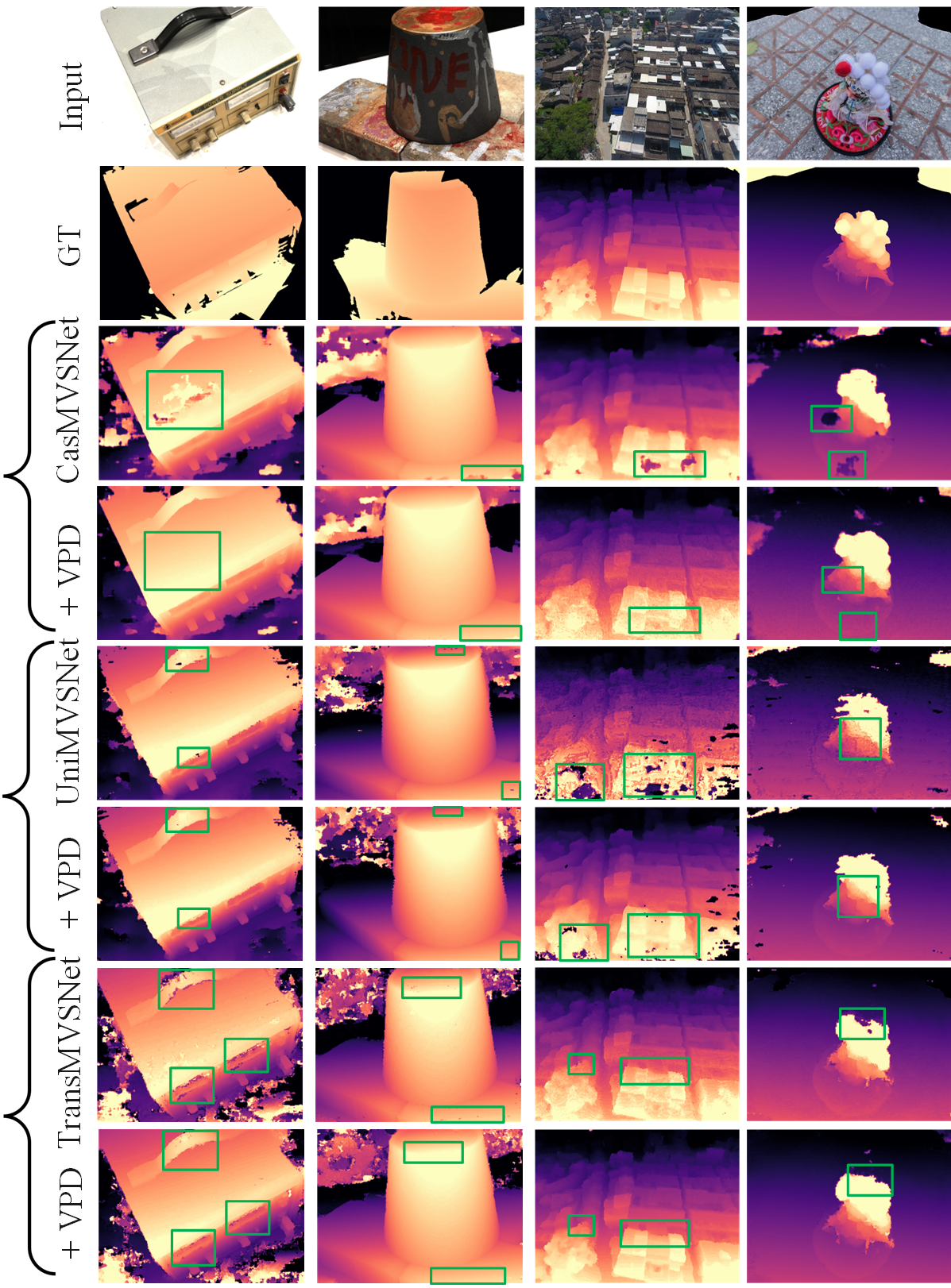}
\caption{
Qualitative results for MVS on DTU test set (left two columns) and BlendedMVS validation set (right two columns).
% Different baseline networks are employed to evaluate the effectiveness of our proposed method. 
% From top to bottom,
% (a) reference images, (b) ground truth, (c) CasMVSNet~\cite{gu2020cascade} results, (d) CasMVSNet~\cite{gu2020cascade} + VPD results, (e) UniMVSNet~\cite{peng2022rethinking} results, (f) UniMVSNet~\cite{peng2022rethinking} + VPD results, (g) TransMVSNet~\cite{ding2022transmvsnet} results, (h) TransMVSNet~\cite{ding2022transmvsnet} + VPD results. 
Our approach consistently generates more complete predictions in low-texture regions,
% (e.g., bottom regions in column 3, and the middle region of the object in column 4), 
as well as more accurate and fine-grained results in thin-structure regions.
% (e.g., the edges of the object in column 1).
}
\label{dtu_blend}
\end{figure}

\begin{table*}[!ht] 
\begin{center}
\scriptsize
\begin{tabular}{l|cccccccccc}
\toprule
 Method & Abs Rel $\downarrow$ & Abs $\downarrow$ & Sq Rel $\downarrow$ & Th8 $\downarrow$ & Th20 $\downarrow$ & $\delta_{1}<1.25 $ $\uparrow$ & $\delta_{2}<1.25^{2}$  $\uparrow$   \\ \midrule
MVSNet~\cite{yao2018mvsnet} & 0.0139 & 11.5502 & 2.0383  &  0.1378 & 0.0932 & 0.9845 & 0.9966    \\
CasMVSNet~\cite{gu2020cascade} & 0.0097 & 7.4381 & 1.6300 & 0.0872 & 0.0570 & 0.9887 & 0.9976   \\
UniMVSNet~\cite{peng2022rethinking} & 0.0095 &  7.2756 & 1.3163 & 0.0837 & 0.0547 & 0.9858 &  0.9934    \\
TransMVSNet~\cite{ding2022transmvsnet} & 0.0094 & 7.2096 & 1.2712   & 0.0842 & 0.0541 & 0.9905 & 0.9982   \\ 
\hline
% \rowcolor{gray!10}
% CasMVSNet~\cite{gu2020cascade}$+$Ours & 0.0075 & 5.7275 &  1.2439  & 0.0644 & 0.0414 & 0.9913 &  0.9981  \\
% \rowcolor{gray!10}
% UniMVSNet~\cite{peng2022rethinking}$+$Ours & 0.0071 & 5.2383 & 1.0402  & 0.0574 & 0.0404 & 0.9870 & 0.9940   \\
\rowcolor{gray!10}
TransMVSNet~\cite{ding2022transmvsnet} $+$ Ours & \textbf{0.0067} & \textbf{4.9416} & \textbf{0.9918}  & \textbf{0.0510} & \textbf{0.0333} & \textbf{0.9918} & \textbf{0.9984}    \\
 \bottomrule
\end{tabular}
\caption{ Quantitative results on DTU test set for MVS. The best performers are marked \textbf{bold}. }
\label{table:dtu}
\end{center}
\end{table*}

\begin{table}[!ht] 
\begin{center}
\scriptsize
\begin{tabular}{l|cccc}
\toprule
 Method & Abs Rel $\downarrow$ & Abs $\downarrow$  & $\delta_{1}<1.25$ $\uparrow$\\ \midrule
MVSNet~\cite{yao2018mvsnet} & 0.0915 & 2.6554 & 0.9135  \\
CasMVSNet~\cite{gu2020cascade} & 0.0665 &  1.7102  & 0.9349 \\
UniMVSNet~\cite{peng2022rethinking} &  0.0825 & 1.8744 & 0.9320 \\
TransMVSNet~\cite{ding2022transmvsnet} & 0.0657 & 1.9216  & 0.9402\\ 
\hline
\rowcolor{gray!10}
CasMVSNet~\cite{gu2020cascade} $+$ Ours &  0.0404 & 1.4122  & {0.9604}  \\
\rowcolor{gray!10}
UniMVSNet~\cite{peng2022rethinking} $+$ Ours & 0.0496 & 1.3128 &  \textbf{0.9425}\\
\rowcolor{gray!10}
TransMVSNet~\cite{ding2022transmvsnet} $+$ Ours & \textbf{0.0376} & \textbf{1.2267} & 0.9596 \\
 \bottomrule
\end{tabular}
\caption{ Zero-shot generalization from DTU to BlendedMVS validation set for MVS. }
\label{tabblen}
\end{center}
\end{table}

\begin{table}[!ht] 
\begin{center}
\renewcommand\tabcolsep{1.2pt}
\scriptsize
\begin{tabular}{l|cccccc}
\toprule
  Methods  &  Abs Rel $\downarrow$ & Abs $\downarrow$ & Sq Rel $\downarrow$ &  RMSE $\downarrow$  & $\delta_{1}<1.25$ $\uparrow$  \\ \midrule
MVDepth~\cite{wang2018mvdepthnet} &  0.1167  &  0.2301  &  0.0596  & 0.3236 & 0.8453    \\ 
DPSNet~\cite{im2019dpsnet} & 0.1200   & 0.2104   & 0.0688  & 0.3139  & 0.8640   \\  
DELTAS~\cite{sinha2020depth} & 0.0915   &  0.1710  & 0.0327  & 0.2390  & 0.9147    \\ 
NRGBD~\cite{liu2019neural} & 0.1013   & 0.1657   & 0.0502  & 0.2500  &0.9160    \\
PairNet~\cite{wang2022itermvs} & 0.0895   & 0.1709   & 0.0615  & 0.2734  & 0.9172   \\ \hline
ESTD~\cite{long2021multi} &  0.0812  & 0.1505  & 0.0298  &0.2199   &0.9313      \\  
\rowcolor{gray!10}
ESTD~\cite{long2021multi} $+$ Ours &  \textbf{0.0753}  & \textbf{0.1497}  & \textbf{0.0237}  & \textbf{0.2149}  &  \textbf{0.9483}     \\     \bottomrule
\end{tabular}
\caption{ Quantitative results on ScanNet test set for MVS.\looseness=-1}
\label{tabscan}
\end{center}
\end{table}

\section{Experiments}
We evaluate the proposed {Volumetric Probability Diffusion (VPD)} on the 3D scene perception tasks of multi-view stereo (MVS) and semantic scene completion (SSC).

\subsection{Multi-view Stereo (MVS)}
\subsubsection{Datasets.} {DTU}~\cite{article} is a large-scale indoor dataset, which consists of 124 different scenes with 7 different illumination conditions. We split the dataset into training, validation, and test set following the setting of MVSNet~\cite{yao2018mvsnet}. 
{BlendedMVS}~\cite{yao2020blendedmvs} dataset is a synthetic dataset that consists of 106 training scans and  7 validation scans.   
{ScanNet}~\cite{dai2017scannet} is an RGB-D video dataset that consists of more than 1600 scans, annotated with depth maps and camera poses.

\subsubsection{Implementation Details.} Our model is implemented on the Pytorch platform with 4 NVIDIA A100 GPUs.
We train our model for 16 epochs on the DTU dataset and 7 epochs on the ScanNet dataset, respectively. For the BlendedMVS dataset, we implement tests using the model trained on the DTU dataset to evaluate the generalization ability. The initial learning rate is set to $2.5\times10^{-5}$, which decays following the same strategy of baseline networks.
During the training process, the batch size is set to 12 and we adopt Adam as the optimizer. The diffusion forward step $\boldsymbol{T}$ is set to 1000 and we adopt 4 iterations in the reverse process.

\begin{figure*}[!ht]
	\begin{center}
            \hsize=\textwidth %
		\includegraphics[width=1.0\textwidth]{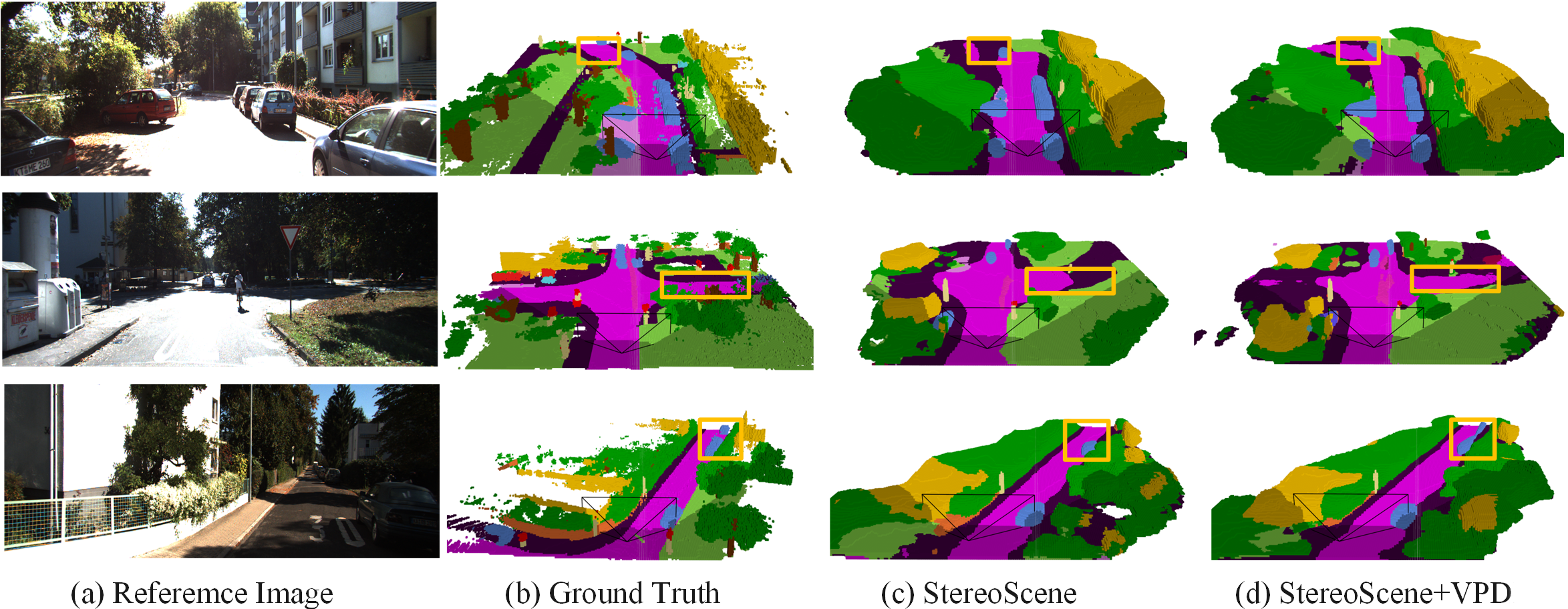}  
		\begin{tabular}{cccccc}
			\multicolumn{6}{c}{
				\scriptsize
				\textcolor{bicycle}{$\blacksquare$}bicycle~
				\textcolor{car}{$\blacksquare$}car~
				\textcolor{motorcycle}{$\blacksquare$}motorcycle~
				\textcolor{truck}{$\blacksquare$}truck~
				\textcolor{other-vehicle}{$\blacksquare$}other vehicle~
				\textcolor{person}{$\blacksquare$}person~
				\textcolor{bicyclist}{$\blacksquare$}bicyclist~
				\textcolor{motorcyclist}{$\blacksquare$}motorcyclist~
				\textcolor{road}{$\blacksquare$}road~
				}
    \\
    
			\multicolumn{6}{c}{
				\scriptsize
                    \textcolor{parking}{$\blacksquare$}parking~
				\textcolor{sidewalk}{$\blacksquare$}sidewalk~
				\textcolor{other-ground}{$\blacksquare$}other ground~
				\textcolor{building}{$\blacksquare$}building~
				\textcolor{fence}{$\blacksquare$}fence~
				\textcolor{vegetation}{$\blacksquare$}vegetation~
				\textcolor{trunk}{$\blacksquare$}trunk~
				\textcolor{terrain}{$\blacksquare$}terrain~
				\textcolor{pole}{$\blacksquare$}pole~
				\textcolor{traffic-sign}{$\blacksquare$}traffic sign			
			}
		\end{tabular}	
 	\end{center}
\caption{
Qualitative results for SSC on SemanticKITTI validation set. The shadow areas denote unseen scenery out of the camera’s field of view. Our proposed VPD improves the performance of the baseline in challenging regions. 
% (e.g. textureless areas and small objects).
}
\label{figsk}
\end{figure*}

\subsubsection{Performance.} For quantitative evaluation, we conduct standard metrics~\cite{eigen2014depth,long2021multi,cai2022riav}, including absolute relative error (Abs Rel), absolute error (Abs), square relative error (Sq Rel), root mean square error in linear scale (RMSE), threshold distance error (Th) and inlier ratios ($\delta< 1.25^{i}$, where $i\in \left\{ 1,2 \right\}$).
As reported in Table~\ref{table:dtu}, our method shows significant improvements compared to TransMVSNet, reducing Abs by $31.46\%$.

We evaluate the zero-shot generalization ability of our method from DTU to BlendedMVS validation set without any fine-tuning. As shown in Table~\ref{tabblen}, our method has a notable performance gain compared to baseline networks, which demonstrates that our approach generalizes well across different datasets without post-processing.
Table~\ref{tabscan} shows quantitative results on the ScanNet test set. 
ESTD~\cite{long2021multi} with VPD outperforms other methods in terms of accuracy, which indicates that our method also has strong modeling capability for temporal cost volumes.\looseness=-1

Moreover, We visualize qualitative results on the DTU test set and BlendedMVS validation set in Figure~\ref{dtu_blend}. 
{Our approach significantly enhances outcomes from baseline models, generating more complete depth maps with heightened accuracy, particularly in challenging areas like object boundaries and repetitive patterns.}

\begin{table*}[!ht]
\begin{center}
\scriptsize
\setlength{\tabcolsep}{0.003\linewidth}
\newcommand{\classfreq}[1]{{~\tiny(\semkitfreq{#1}\%)}}  %
\centering
\begin{tabular}{l|c c c c c c c c c c c c c c c c c c c c |c}
    \toprule
    Method & Input
    & \rotatebox{90}{\textcolor{road}{$\blacksquare$} road\classfreq{road}} 
    & \rotatebox{90}{\textcolor{sidewalk}{$\blacksquare$} sidewalk\classfreq{sidewalk}}
    & \rotatebox{90}{\textcolor{parking}{$\blacksquare$} parking\classfreq{parking}} 
    & \rotatebox{90}{\textcolor{other-ground}{$\blacksquare$} other-grnd\classfreq{otherground}} 
    & \rotatebox{90}{\textcolor{building}{$\blacksquare$} building\classfreq{building}} 
    & \rotatebox{90}{\textcolor{car}{$\blacksquare$} car\classfreq{car}} 
    & \rotatebox{90}{\textcolor{truck}{$\blacksquare$} truck\classfreq{truck}} 
    & \rotatebox{90}{\textcolor{bicycle}{$\blacksquare$} bicycle\classfreq{bicycle}} 
    & \rotatebox{90}{\textcolor{motorcycle}{$\blacksquare$} motorcycle\classfreq{motorcycle}} 
    & \rotatebox{90}{\textcolor{other-vehicle}{$\blacksquare$} other-veh.\classfreq{othervehicle}} 
    & \rotatebox{90}{\textcolor{vegetation}{$\blacksquare$} vegetation\classfreq{vegetation}} 
    & \rotatebox{90}{\textcolor{trunk}{$\blacksquare$} trunk\classfreq{trunk}} 
    & \rotatebox{90}{\textcolor{terrain}{$\blacksquare$} terrain\classfreq{terrain}} 
    & \rotatebox{90}{\textcolor{person}{$\blacksquare$} person\classfreq{person}} 
    & \rotatebox{90}{\textcolor{bicyclist}{$\blacksquare$} bicyclist\classfreq{bicyclist}} 
    & \rotatebox{90}{\textcolor{motorcyclist}{$\blacksquare$} motorcyclist.\classfreq{motorcyclist}} 
    & \rotatebox{90}{\textcolor{fence}{$\blacksquare$} fence\classfreq{fence}} 
    & \rotatebox{90}{\textcolor{pole}{$\blacksquare$} pole\classfreq{pole}} 
    & \rotatebox{90}{\textcolor{traffic-sign}{$\blacksquare$} traf.-sign\classfreq{trafficsign}}  & mIoU
    \\
    \midrule
    MonoScene~\cite{cao2022monoscene} & Mono  & 54.70 & 27.10 & 24.80 & {5.70} & 14.40 & 18.80 & 3.30 & 0.50 & 0.70 & 4.40 & 14.90 & 2.40 & 19.50 & 1.00 & 1.40 & 0.40 & 11.10 & 3.30 & 2.10 & 11.08 \\
    VoxFormer-S~\cite{li2023voxformer} & Stereo  & 53.90 & 25.30 & 21.10 & 5.60 & 19.80 & 20.80 & 3.50 & 1.00 & 0.70 & 3.70 & 22.40 & 7.50 & 21.30 & 1.40 & \textbf{2.60} & 0.20 & 11.10 & 5.10 & 4.90 & 12.20 \\
    VoxFormer-T~\cite{li2023voxformer} & Stereo-T & 54.10 & 26.90 & 25.10 & 7.30 & 23.50 & 21.70 & 3.60 &  1.90 & 1.60 & 4.10 & 24.40 & 8.10 & 24.20 & 1.60 & 1.10 & 0.00 & 13.10 & 6.60 & 5.70 & 13.41 \\
    SSCNet~\cite{song2017semantic}  &  LiDAR & 51.15 & 30.76 & 27.12 & 6.44 & \textbf{34.53} & 24.26& 1.18 &0.54 &{0.78}  &4.43 &\textbf{35.25} &1.18 &29.01 &0.25 &0.25 &\textbf{0.78} &\textbf{19.87} &\textbf{13.10} &6.73  & 16.14\\ \hline
    
    StereoScene~\cite{li2023stereoscene} & Stereo & \textbf{61.90} & 31.20 & \textbf{30.70} & 10.70 & 24.20 & 22.80 & {2.80} & {3.40} & {2.40} & 6.10 & 23.80 & 8.40 & 27.00 & \textbf{2.90} & {2.20} & 0.50 & {16.50} & 7.00 & 7.20 & 15.36 \\
    \rowcolor{gray!10} StereoScene~\cite{li2023stereoscene} $+$ Ours & Stereo & {61.76} & \textbf{32.41} & {20.39} & \textbf{11.11} & {24.43} & \textbf{32.12} & \textbf{7.33} & \textbf{3.74} & \textbf{2.53} & \textbf{9.26} & {25.87} & \textbf{8.89} & \textbf{37.68} & 2.02 & {0.99} & 0.00 & 10.09 & {11.72} & \textbf{7.53} & \textbf{16.31} \\
    \bottomrule
    \end{tabular}
    \caption{ Quantitative results on SemanticKITTI test set against the state-of-the-art SSC methods (higher is better). Our method even surpasses \textbf{temporal stereo-based} (Stereo-T) VoxFormer-T and \textbf{LiDAR-based} SSCNet in terms of mIoU.}
\label{tabsk}
\end{center}
\end{table*}

\subsection{Semantic Scene Completion (SSC)}

\subsubsection{Datasets.} SemanticKITTI~\cite{behley2019semantickitti} is a popular semantic scene completion dataset, which contains 22 outdoor driving scenes. SemanticKITTI holds LiDAR annotations that are voxelized as 256$\times$256$\times$32 grid of 0.2m voxels. The target voxel grids are labeled with 21 classes (1 free, 1 unknown, and 19 semantics).
Following~\cite{li2023stereoscene}, we only adopt RGB images of the dataset as inputs.

\subsubsection{Implementation Details.} We extend StereoScene~\cite{li2023stereoscene} with our proposed VPD for SSC evaluation. More specifically, the geometric cost volume in StereoScene is leveraged as the coarse volume.
The whole model is trained on SemanticKITTI for 30 epochs with a learning rate of $2.5\times10^{-5}$. AdamW is adopted as a training optimizer 
following~\cite{li2023stereoscene}.

\subsubsection{Performance.} For quantitative evaluation, we adopt mIoU (mean Intersection over Union) to account for the SSC task. We compare our method with other state-of-the-art SSC networks: (1) camera-based methods including  MonoScene~\cite{cao2022monoscene}, 
VoxFormer~\cite{li2023voxformer}
% VoxFormer-S~\cite{li2023voxformer}, VoxFormer-T~\cite{li2023voxformer} 
and StereoScene~\cite{li2023stereoscene}, (2) LiDAR-based method of SSCNet~\cite{song2017semantic}.
As shown in Table~\ref{tabsk}, our method surpasses StereoScene by $6.18\%$ in terms of mIoU, which demonstrates the application of VPD effectively improves the accuracy of depth estimation and thereby enhances the performance of semantic scene completion.
It's worth noting that our method even surpasses the LiDAR-based method of SSCNet in terms of mIoU.
Figure~\ref{figsk} visualizes the qualitative results, our method produces more accurate and complete results in large-scale driving scenarios compared with StereoScene.

\begin{table}
\begin{center}
% \renewcommand\tabcolsep{2.0pt}
% \captionsetup{type=table}
\scriptsize
\begin{tabular}{ccc|cc}
\toprule
\multicolumn{3}{c|}{\textbf{Model Settings}} & \multicolumn{2}{c}{\textbf{Evaluation Metrics}} \\ \midrule
CVP & CACC & OF & Abs Rel $\downarrow$ & Abs $\downarrow$ \\ \hline
  &  &  & 0.0097 & 7.4381  \\
 \checkmark  &  &       &  0.0083&   6.3111 \\ 
 \checkmark &   \checkmark &   &  0.0078 & 5.9829   \\
 \checkmark &   \checkmark & \checkmark  & \textbf{0.0075} &  \textbf{5.7275}  \\  \bottomrule
\end{tabular}
\caption{Ablation study for different model settings. The basic CasMVSNet is employed as the baseline setting.
}
\label{ablation}
\end{center}
\end{table}

\subsection{Ablation Study}\label{sec43}

We conduct extensive ablation studies on DTU test set with different model settings. We extend CasMVSNet with VPD of different settings for the evaluation.

\subsubsection{Effect of Model Settings.} For the experiment in the first row, we adopt basic CasMVSNet as the baseline setting. The reverse iterations are set to 4 unless otherwise stated.
As shown in Table~\ref{ablation}, the VPD framework brings significant improvement with the basic prior volume condition of CVP. 
The CACC and OF obviously enhance the depth estimation performance with reliable volumetric learning, reducing Abs by $5.20\%$ and $4.27\%$, respectively.

\begin{table}[!ht]
\begin{center}
\scriptsize
\renewcommand\tabcolsep{0.8pt}
\begin{tabular}{l|ccc}
\toprule
 {Method}& Abs $\downarrow$ & Run-time(s) $\downarrow$ & Memory(G) $\downarrow$ \\  \midrule
MVSNet~\cite{yao2018mvsnet}&	11.55&	0.75&	12.74      \\
CasMVSNet~\cite{gu2020cascade}& 7.44&	\textbf{0.41}&	\textbf{6.32}    \\
\rowcolor{gray!10}CasMVSNet~\cite{gu2020cascade} + Ours (4 Steps)&	5.73& 0.69& 8.11    \\ 
TransMVSNet~\cite{ding2022transmvsnet}&	7.21&	0.87&	8.27    \\ 
TransMVSNet~\cite{ding2022transmvsnet} + Ours (1 Step)&	6.27& 0.96&	10.48 \\  
TransMVSNet~\cite{ding2022transmvsnet} + Ours (2 Steps)& 5.40& 1.04& 10.48\\
\rowcolor{gray!10}TransMVSNet~\cite{ding2022transmvsnet} + Ours (4 Steps)& 4.94& 1.24& 10.48\\
TransMVSNet~\cite{ding2022transmvsnet} + Ours (6 Steps)& 4.79& 1.39& 10.48\\
TransMVSNet~\cite{ding2022transmvsnet} + Ours (8 Steps)& \textbf{4.72}& 1.52& 10.48\\
  \bottomrule
\end{tabular}
\caption{ Running time and memory consumption of the proposed method. We adopt 4 reverse steps as the default setting to balance efficiency and effectiveness. }
\label{efficiency}
\end{center}
\end{table}

\subsubsection{Efficiency Analyse.}
We report the running time and memory consumption of several schemes that are built upon different baselines equipped with our proposed VPD on the NVIDIA A100 GPU, which are detailed in Table~\ref{efficiency}. It's worth noting that our method could effectively achieve compelling performance gains with acceptable time consumption. As shown in the table, the proposed VPD delivers satisfactory performance improvements with relatively slight increments in time consumption, while the memory consumption of our proposed method remains constant regardless of the number of reverse steps.
In addition, the performance gain of more than 4 reverse steps with more running time is not obvious, thus we adopt 4 steps as the default setting to balance efficiency and effectiveness.

\section{Conclusion}
In this work, we propose a novel framework of Volumetric Probability Diffusion (VPD) for 3D scene perception tasks including MVS and SSC. Different from previous single-step approximation solutions, we employ multi-step generative diffusion to progressively model volumetric probability for more reliable estimation.
Specifically, we introduce a Confidence-Aware Contextual Collaboration (CACC) module to correct the uncertain regions for reliable target volume approximation. 
In the sampling process, we develop an Online Filtering (OF) strategy to maintain consistency in estimated volume representations.
Our method achieves state-of-the-art performance on multiple MVS/SSC benchmarks and even surpasses the LiDAR-based method with only camera-based inputs.\looseness=-1

\section{Acknowledgements}

{This paper is supported in part by NSFC under Grant 62302246 and ZJNSFC under Grant LQ23F010008.}

\appendix
\twocolumn[{
\centering
 \vspace{20pt}
\section*{\Large \centering Supplementary Material of VPD}
 \vspace{30pt}
 }]

\section{Implementation of Task-Specific Head}
We report the implementation details for the task-specific head as follows.

For multi-view stereo (MVS), the task-specific head is constructed following different baseline networks~\cite{gu2020cascade,peng2022rethinking,ding2022transmvsnet,long2021multi}. Specifically, in CasMVSNet~\cite{gu2020cascade} and ESTD~\cite{long2021multi}, the estimated depth map is computed with differentiable soft-argmin operation along the depth direction of the estimated volume. In UniMVSNet~\cite{peng2022rethinking}, unity regression is employed to regress the depth map, which selects the optimal hypothesis and calculates the offset to ground-truth depth. TransMVSNet~\cite{ding2022transmvsnet} leverages the Winner-Takes-All (WTA)~\cite{cheng2020hierarchical} operation along the depth dimension to get the estimated depth map.

For semantic scene completion (SSC), the estimated volume will be upsampled and processed with a softmax layer to get the semantic occupancy prediction following StereoScene~\cite{li2023stereoscene}.

\section{Architectural Details on Diffusion Unet}
As described in Section 3 of the main paper, we adopt a 3D UNet~\cite{ronneberger2015u,muller2022diffrf} as the architecture of the volumetric diffusion model. Specifically, the 3D UNet consists of three downsample and upsample blocks. 
Each downsample block contains two residual layers~\cite{he2016deep} and a CACC module for the contextual feature condition guidance, while each upsample block contains two residual layers.
The CACC takes contextual features as inputs to refine the estimated depth volumes in the 3D UNet. 
The downsample blocks adopt 3D convolutions with stride two for downsampling, and the upsample blocks use trilinear interpolation for upsampling.
We employ Group Norm~\cite{wu2018group} and SiLU~\cite{elfwing2018sigmoid} in each block for normalization and activation, respectively.
Moreover, the basic prior volume condition is generated from the given cost volume, which is concatenated with the input noisy volume before feeding into the 3D UNet.

\begin{figure}[!ht]
\vspace{-0pt}
\centering
\includegraphics[width=1.0\linewidth]{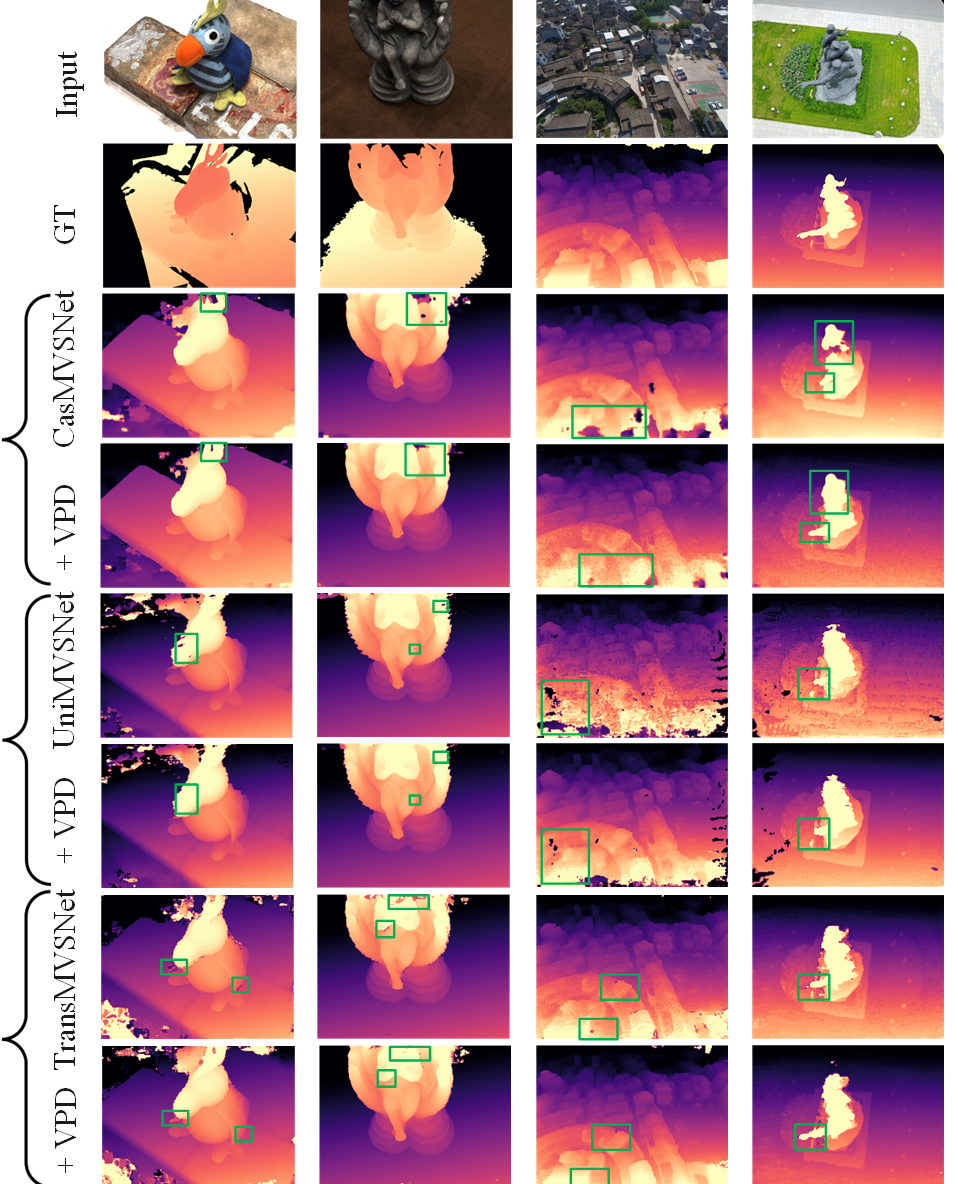}
\caption{
More visualization results for MVS on the DTU~\cite{article} test set (left two columns) and the BlendedMVS~\cite{yao2020blendedmvs} validation set (right two columns). Our method generates more complete and accurate results in challenging regions.}
\label{supple_mvs}
 \vspace{-0pt}
\end{figure}

\begin{figure*}
   \vspace{-10pt}
	\begin{center}
            \hsize=\textwidth %
		\includegraphics[width=0.9\textwidth]{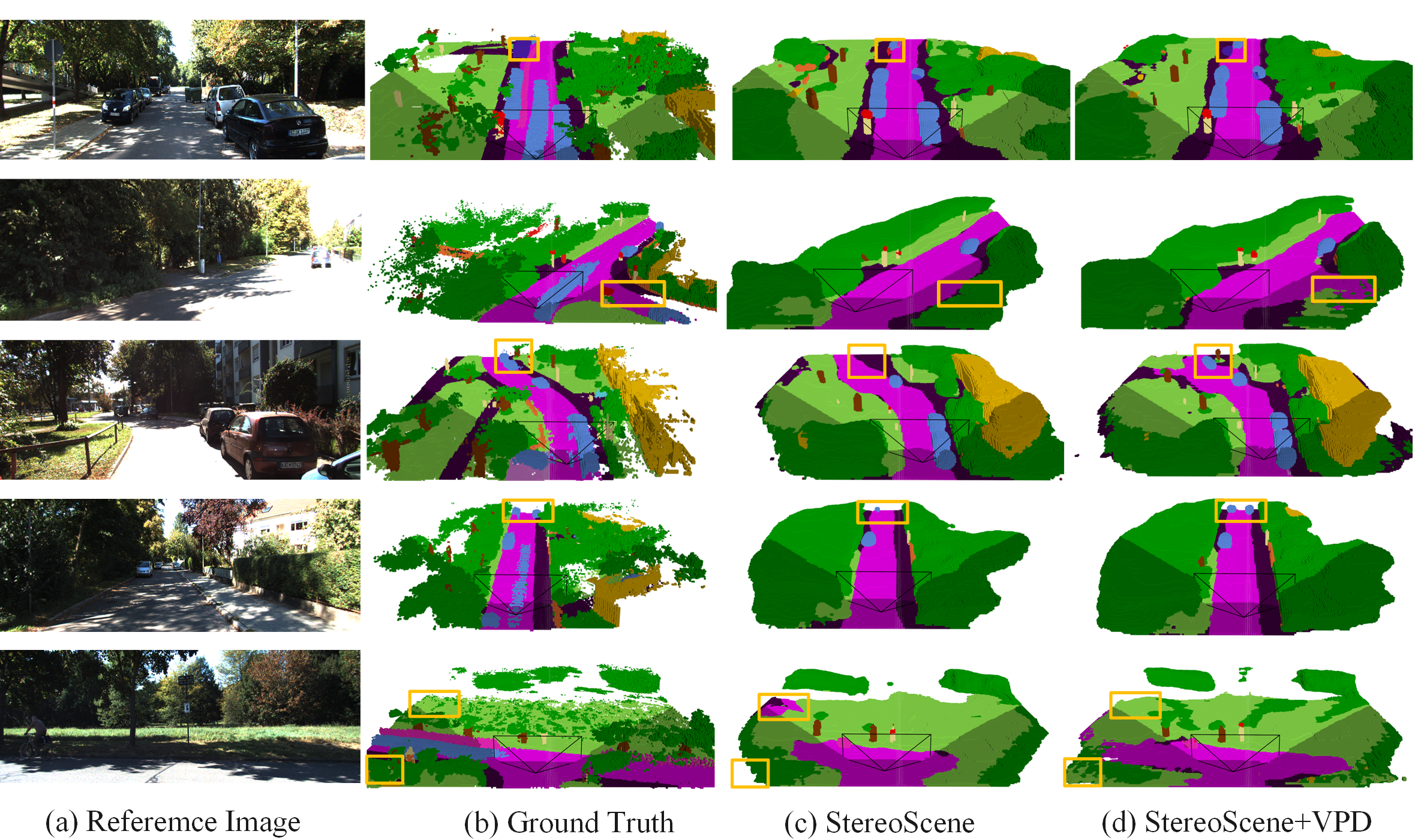}  
		\begin{tabular}{cccccc}
			\multicolumn{6}{c}{
				\scriptsize
				\textcolor{bicycle}{$\blacksquare$}bicycle~
				\textcolor{car}{$\blacksquare$}car~
				\textcolor{motorcycle}{$\blacksquare$}motorcycle~
				\textcolor{truck}{$\blacksquare$}truck~
				\textcolor{other-vehicle}{$\blacksquare$}other vehicle~
				\textcolor{person}{$\blacksquare$}person~
				\textcolor{bicyclist}{$\blacksquare$}bicyclist~
				\textcolor{motorcyclist}{$\blacksquare$}motorcyclist~
				\textcolor{road}{$\blacksquare$}road~
				}
    \\
    
			\multicolumn{6}{c}{
				\scriptsize
                    \textcolor{parking}{$\blacksquare$}parking~
				\textcolor{sidewalk}{$\blacksquare$}sidewalk~
				\textcolor{other-ground}{$\blacksquare$}other ground~
				\textcolor{building}{$\blacksquare$}building~
				\textcolor{fence}{$\blacksquare$}fence~
				\textcolor{vegetation}{$\blacksquare$}vegetation~
				\textcolor{trunk}{$\blacksquare$}trunk~
				\textcolor{terrain}{$\blacksquare$}terrain~
				\textcolor{pole}{$\blacksquare$}pole~
				\textcolor{traffic-sign}{$\blacksquare$}traffic sign			
			}
		\end{tabular}	
 	\end{center}
        \vspace{-10pt}
\caption{
More visualization results on the SemanticKITTI~\cite{behley2021towards} validation set. The shadow areas denote unseen scenery out of the camera’s field of view.  Our method generates more accurate 3D scene layouts and shows obvious advancement on small moving objects.
}
\label{supple_ssc}
\end{figure*}

\section{Additional Qualitative Results}
We provide additional qualitative results for MVS and SSC. The results further demonstrate the effectiveness of our approach in enhancing the 3D scene perception performance.

\textbf{Multi-view Stereo (MVS). }We visualize more qualitative results on the DTU~\cite{article} test set and the BlendedMVS~\cite{yao2020blendedmvs} validation set in Figure~\ref{supple_mvs}. 
For the tests on the BlendedMVS dataset, we use the model trained on the DTU dataset to evaluate the generalization ability.
As described in Section 3 of the main paper, the initial coarse prior the VPD framework is continuously improved with the proposed CACC module and OF strategy during the diffusion process.
Therefore, compared with the baseline networks, our method generates more complete depth maps and achieves more accurate results in challenging regions (e.g., object boundaries in column 1, repetitive pattern regions in column 2 and column 4).

\textbf{Semantic Scene Completion (SSC). }We visualize more qualitative results on the SemanticKITTI validation set in Figure~\ref{supple_ssc}. 
Compared to StereoScene~\cite{li2023stereoscene}, our method shows obvious advancement on small moving objects (e.g., trucks in row 1, cars in row 3 and row 4) and generates more accurate 3D scene layout (e.g., roads in row 2 and row 5).

\section{Evaluation of CACC and OF on Baselines}

We conduct additional experiments by directly applying CACC and OF to the cost volume of CasMVSNet as shown in Table~\ref{table_2}. The improvements of such implementation are obviously less compared with applying CACC and OF to our proposed framework, which we attribute to the fact that CACC and OF are optimally functional only within our multi-step generative process.

\begin{table}[!ht] 
\begin{center}
\scriptsize
\begin{tabular}{l|c|c}
\toprule
Method & Abs Rel $\downarrow$   & Abs $\downarrow$   \\ \midrule
CasMVSNet&	0.0097&	7.4381 \\
CasMVSNet+CACC+OF&	0.0091&	7.1218 \\
\rowcolor{gray!10}CasMVSNet+VPD  &  \textbf{0.0075}	&\textbf{5.7275}   \\
\bottomrule
\end{tabular}
\caption{ Evaluation results of applying CACC and OF to the cost volume of the baseline network. }
\vspace{0pt}
\label{table_2}
\end{center}
\end{table}

% \section{Limitations and Ethical Concerns}
% The incomplete ground truth labeling in the datasets could affect the training of the proposed method. 

% While the promising deep prediction results of the proposed method could promote the development of autonomous driving, the legal challenges, as well as the privacy and data security risks of autonomous driving remain subjects of debate.

\bibliography{ref}

\end{document}